\pdfoutput=1
\documentclass[11pt]{article}

\usepackage{acl}
\usepackage{nert}

\usepackage{pgfplots}
\pgfplotsset{compat=1.17}

\renewcommand{\nertcomment}[4]{\unskip}

\usepackage{times}
\usepackage{latexsym}
\usepackage{graphicx}
\graphicspath{ {./images/} }
\usepackage{linguex}
\usepackage{tabularx}
\usepackage[utf8]{inputenc}

\newcommand{\sem}[1]{\textsc{#1}}

\title{Construction Identification and Disambiguation Using BERT:\\ A Case Study of NPN}

\author{Wesley Scivetti \quad Nathan Schneider \\
  Georgetown University \\
  \{\emldisplay{wss37@georgetown.edu}{wss37}, \emldisplay{nathan.schneider@georgetown.edu}{nathan.schneider}\}\texttt{@georgetown.edu}}
  
\date{}

\begin{document}
\maketitle
\begin{abstract}

%In recent years, there has been much research on probing Language Models (LMs) in order to judge their linguistic knowledge and capabilities. While there is a wealth of research indicating models of various sizes are able to correctly learn syntactic and semantic generalizations in language, there remain 

%Despite research on linguistic probing of Language Models (LMs) indicating models of various sizes are able to correctly learn syntactic and semantic generalizations in language,
%a broad range of linguistic phenomena remain underexplored. From the perspective of linguistic theory, Construction Grammar provides an elegant account of many such phenomena which are often discounted as part of the ``periphery'' of language. 

Construction Grammar hypothesizes that knowledge of a language consists chiefly of knowledge of form--meaning pairs (``constructions'') that include vocabulary, general grammar rules, and even idiosyncratic patterns.
Recent work has shown that transformer language models 
%have the capability to recognize and understand 
represent at least some constructional patterns, including ones where the construction is rare overall. In this work, we 
probe BERT's representation of the form and meaning of a minor construction of English, the NPN (noun--preposition--noun) construction---exhibited in such expressions as \textit{face to face} and \textit{day to day}---which is known to be polysemous.
We construct a benchmark dataset of semantically annotated corpus instances (including distractors that superficially resemble the construction). With this dataset, we train and evaluate probing classifiers. 
They achieve decent discrimination of the construction from distractors, as well as sense disambiguation among true instances of the construction, revealing that BERT embeddings carry indications of the construction's semantics.
Moreover, artificially permuting the word order of true construction instances causes them to be rejected, indicating sensitivity to matters of form.
%
%test if BERT can identify a construction which has not been studied in previous LM research, namely the NPN construction. Furthermore, we introduce the task of construction sense disambiguation, and test if the model can determine the correct constructional semantics in context. We find that a linear probe over BERT representations is able to distinguish cases of the construction from near minimal pairs from corpus data, and is robust to artificial word order changes. Similarly, we find that BERT is able to disambiguate between multiple core meanings of the construction beyond what is possible using a non-contextual lexical semantic baseline. 
%
We conclude that BERT does latently encode at least some knowledge of the NPN construction going beyond a surface syntactic pattern and lexical cues.\footnote{Code and Data: \url{https://github.com/WesScivetti/NPN_probing}}

\end{abstract}

\section{Introduction}

The ``black box'' nature of Language Models (LMs) like has spawned a great deal of research investigating the extent to which these LMs are able to represent and understand a variety of linguistic phenomena \citep{linzenSyntacticStructureDeep2021a,Rogers_Kovaleva_Rumshisky_2021,Chang_Bergen_2024}. There has been substantial work focusing on many aspects of linguistic knowledge, including hierarchical structure \citep{Clark_Khandelwal_Levy_Manning_2019, Hewitt_Manning_2019,Jawahar_Sagot_Seddah_2019}, lexical semantics \citep{Chang_Chen_2019,Vulić_Ponti_Litschko_Glavaš_Korhonen_2020}, negation \citep{Ettinger_2020}, agreement phenomena \citep{Linzen_Dupoux_Goldberg_2016,weissweilerCountingBugsChatGPT`s2023a}, and filler-gap dependencies \citep{Wilcox_Levy_Morita_Futrell_2018,wilcoxUsingComputationalModels2024}. Broadly, these results show that even relatively modest sized LSTMs and transformer models are able to demonstrate nontrivial (though far from perfect) linguistic knowledge. However, there is some indication that these models are sometimes reliant on more surface level heuristics, and fail in situations which are straightforward to humans \citep{McCoy_Pavlick_Linzen_2019,Ettinger_2020}. More generally, language models have been generally shown to struggle in out-of-domain situations  \citep{mccoyEmbersAutoregressionShow2024} and have some difficulty applying linguistic paradigms to nonce words \citep{weissweilerCountingBugsChatGPT`s2023a} and rare syntactic constructions \citep{Scivetti_Torgbi_Blodgett_Shichman_Hudson_Bonial_Madabushi_2025}.

Thus, there is need to evaluate language models on a range of linguistic tasks which go beyond the more studied ``core'' linguistic phenomena. Such work serves to provide a more complete picture of how language models succeed and fail across the broad spectrum of phenomena in language. Indeed, beyond the more mainstream notions of linguistic structure and information, there is also work on investigating LM knowledge of more idiosyncratic \textit{constructions}, as defined by Construction Grammar. Construction Grammar is broadly a family of linguistic theories which consider all parts of language to be made up of constructions, which are pairings of linguistic \textit{forms} with \textit{meaning} or function (\citealt{Goldberg_1995,Croft_2001}, \textit{inter alia}). It remains unclear the extent to which LMs may implicitly view constructions as distinct units. Because of their emphasis on pairing form with meaning, CxG theories provide possibilities for testing language model capabilities at the interface of form and meaning for different aspects of language, in contrast to past work which has focused on either syntax (e.g.~\citealt{Hewitt_Manning_2019}) or semantics (e.g.~\citealt{Vulić_Ponti_Litschko_Glavaš_Korhonen_2020}) in isolation. 
\nss{this is introducing a lot of terminology. form and meaning are important. can schematicity/substantivity/entrechment be moved to \S2? are they all needed for the narrative?}\ws{I agree, I took that technical part out.}
A substantial and growing amount of research has recently focused on the intersection of LM knowledge and Construction Grammar (\citealp{Tayyar_Madabushi_Romain_Divjak_Milin_2020,Tseng_Shih_Chen_Chou_Ku_Hsieh_2022,Pannitto_Herbelot_2023,Veenboer_Bloem_2023}, \textit{inter alia}), with a particular focus on argument structure constructions \citep{Li_Zhu_Thomas_Rudzicz_Xu_2022}, the English Comparative Correlative \citep{Weissweiler_Hofmann_Köksal_Schütze_2022}, and the English AANN construction \citep{Chronis_Mahowald_Erk_2023, Mahowald_2023}. While these studies have provided valuable insight into LM processing of constructions with varying levels of schematicity, there remain many constructions which have not been addressed at all in previous work. Furthermore, while \citet{Zhou_Weissweiler_He_Schütze_Mortensen_Levin_2024} do test model understanding of constructions which are similar in form, no past work has focused on individual constructions as polysemous units. We argue this is a gap in past work, as constructions, like words, can have related but distinct meanings that must be properly disambiguated in context in order for correct interpretation. We address this gap by providing experiments which pair formal sensitivity with semantic disambiguation in a controlled manner for a single construction.

\nss{to what extent have those examined meaning/polysemy? if that is a key difference of our approach it's worth saying here}\ws{Addressed}

% These frameworks typically assume that constructions can have varying levels of \textit{schematicity} and \textit{substantivity} with fully substantive constructions (e.g. idioms), and fully schematic constructions (e.g. argument structure constructions). These constructions are hypothesized to become \textit{entrenched} in the mind through a combination of both frequency and idiosyncraticity of the construction \citep{Goldberg_2006}. Due to the continuum of schematic vs.~substantive constructions, each language has a huge number of candidate constructions which have the potential to be entrenched by speakers of a language. 

This work is the first to study whether language models capture the NPN construction \citep{Jackendoff_2008}, an infrequent yet productive pattern exhibited in expressions like \textit{face to face} and \textit{day to day}. 
%Thus, the contributions of this work are to provide a first analysis of how language models handle the NPN construction. Additionally, 
Even for the subset where two instances of the same noun are linked by the preposition \textit{to}, the pattern is polysemous, and sequences matching this pattern on the surface are not always instances of the construction (\cref{sec:npn}).
Guided by CxG theory, we separate our inquiry in terms of the construction's \textit{form} and \textit{meaning} in context. To investigate language modeling of NPN, we:
\begin{itemize}
    \item Construct and annotate a novel dataset of natural NPN examples from COCA (\cref{sec:dataset}).
    \item Probe BERT's ability to distinguish true constructional instances from related constructions and artificial orders (\cref{sec:exp1,sec:perturb}).
    \item Introduce the task of construction sense disambiguation and perform experiments using our dataset (\cref{sec:exp1_semantic}).
\end{itemize}

%\nss{there's a gap here: say we do XYZ (e.g. we construct a novel dataset and probe BERT with classifiers...) good if you can parenthetically reference the sections}\ws{Addressed by creating bulleted list of contributions and then shortening subsequent paragraph}

To summarize our findings, we show that probes using BERT embeddings are able to both identify correct instances of NPN and disambiguate the construction within context at respectable accuracy. Overall, these findings indicate that BERT latently encodes relevant information to the NPN construction, leading to strong sensitivity to both the construction's form and its meaning. 

\nss{it is common to have 3-4 bullet points of key contributions. I think the dataset is one of these. I wonder if the previous paragraph can be replaced. Not all the specific findings need to be articulated in the intro---that is for the conclusion}\ws{Shortened}

\section{The NPN Construction}\label{sec:npn}

\nss{cite UCxn in this section, not just in appendix}

The NPN construction \citep{Jackendoff_2008} follows the general pattern of Noun + Preposition + Noun. Below are 2 examples of the NPN construction. These examples, along with all others, are taken from the Corpus of Contemporary American English (COCA, \citealt{Davies_2010}).

\ex.\label{ex:first} There is a rebellious quality to your
\textbf{day to day}	responses which have not gone unnoticed.

\ex.\label{ex:second} I need you to get this
\textbf{word for word}.

Given the general rules of English, the NPN construction has several unique properties, which we argue separate it from more ``core'' linguistic phenomena. Firstly, the nouns almost always lack determiners, which is unusual for count nouns like ``day''. Secondly, the construction can occur in a variety of syntactic positions, including as an adverbial modifier (as in \cref{ex:second}) and as a prenominal modifier (as in \cref{ex:first}). Finally, it conveys a meaning which is not entirely predictable from its components, and varies considerably depending on the preposition. Common meanings of the NPN construction are the \sem{succession} meaning (shown in \cref{ex:first}) and the \sem{matching/comparison} meaning (shown in \cref{ex:second}). See \citet{Jackendoff_2008} for an overview of the NPN construction and the common meanings associated with various prepositional lemmas. 

While it is conceptually and intuitively appealing to think of NPN as a single construction, some work has argued in favor of viewing NPN as a group of related constructions, which are linked within the mind but not necessarily dominated by a single overarching abstract NPN construction \citep{Sommerer_Baumann_2021}. Due to the wide variety of meanings and distributions of the different NPN constructions, we choose to limit our focus to a single subtype of NPNs, which all share the lemma ``to'' as their preposition, which we refer to as the N\textit{to}N construction. There is still considerable semantic variation even within the N\textit{to}N construction, with 2 broad meanings that we highlight: \sem{succession} (shown in \cref{ex:third}) and  \sem{juxtaposition} (shown in \cref{ex:fourth}). 

\ex.\label{ex:third} I was living \textbf{moment to moment}.

\ex.\label{ex:fourth} You can preserve core warmth by huddling with a buddy, \textbf{chest to chest}.

While there are additional meanings of NPN that do not occur with ``to'' as the preposition, it is one of the only prepositions that is ambiguous in the NPN construction. By not considering examples of NPN with other prepositions, we remove the prepositional lemma as a potential shallow cue that models could learn to predict the construction's semantics. While there are arguably examples of NPNs where the two nouns are not identical, we limit our analysis to cases where the two nouns in the construction match exactly. This allows us to easily gather examples of the construction from corpus data.

\section{Dataset}\label{sec:dataset}

\subsection{Corpus Gathering and Cleaning}
In this work, we endeavor to use natural corpus data to the extent that it was possible. First, we use a simple pattern matching query to extract instances of the sequence Noun + ``to'' + Noun from COCA. We extract the examples from the corpus in a fixed window of +/- 50 tokens from the construction, and then used Stanza \citep{Qi_Zhang_Zhang_Bolton_Manning_2020} to segment the results into sentences and extract the sentences which contained N\textit{to}Ns. We automatically exclude sentences which contained ``from'' preceding the construction, because \textit{from} N \textit{to} N does not have exactly the same distribution as the more general N\textit{to}N \citep{Jackendoff_2008}, and is sometimes studied as a separate (but closely related) construction \citep{Zwarts_2013}. 

After extracting all sentences which contained a possible instance of N\textit{to}N, we then manually clean the data, removing sentences that were either too short (<5 tokens) or contained too many typos. We annotate all instances of the construction for their semantic subtype, and double annotate roughly 25\% of the dataset, achieving an agreement of 84\% and a Cohen's kappa value of .754 between the two annotators, indicating strong agreement. \footnote{Disagreements between the two annotators were resolved through discussion and a gold label was chosen jointly.} The final dataset has 6599 instances of N\textit{to}N, of which 1885 were double annotated.

\subsection{Near Minimal Pairs}
In addition to true instances of the N\textit{to}N construction, we also find grammatical corpus instances of Noun + ``to'' + Noun patterns, which are not instances of the construction. These patterns often occur when a verb licenses a direct object and a ``to'' prepositional phrase, and the direct object and the object of the preposition happen to have the same lemma. Three examples are shown below in \cref{ex:sixth}, \cref{ex:seventh}, and \cref{ex:eight}.

\ex.\label{ex:sixth} Then there's the problem of sticking plastic to plastic.

\ex.\label{ex:seventh} In Rome largesse was doled out by individuals to individuals.

\ex.\label{ex:eight} I don't have time to time travel ...

We do not consider such cases to be examples of the N\textit{to}N construction because the surface pattern of Noun + Preposition + Noun clearly arises from a different syntactic context (e.g.~a verb licensing a direct object and a PP modifier). Furthermore, the meanings of these examples do not evoke the unique semantics that accompany the N\textit{to}N construction. While these cases are not instances of the N\textit{to}N construction, they do provide a set of negative examples which we can use to probe the model's ability to recognize true N\textit{to}N constructions. Throughout this paper, we refer to this set of examples as instances of the N\textit{to}N \textit{distractors}, since we test of if the model is ``distracted'' by the shallow similarity of the examples to the NPN construction. We refer to true examples of N\textit{to}N as instances of the N\textit{to}N \textit{construction}. Since these N\textit{to}N examples exhibit the same surface form as the N\textit{to}N \textit{construction}, we consider them to be near minimal pairs, following \citet{Weissweiler_Hofmann_Köksal_Schütze_2022} who extract near minimal pairs from corpus data based on part-of-speech patterns. While these sentences inevitably contain more lexical biases than a true minimal pair dataset, they are completely natural, and provide a good comparison point for a construction where creating true minimal pairs is otherwise difficult (because there is no obvious minimal change that can be made to result in a grammatical sentence that is not an example of the construction, similar to the struggles of \citet{Weissweiler_Hofmann_Köksal_Schütze_2022} regarding the Comparative Correlative construction). In total, we collect 456 total instances of N\textit{to}N \textit{distractors} from COCA.

\subsection{Train/Test Split}\label{sec:tt_split}

\begin{table}[]
    \centering\small
    \begin{tabular}{cccc}
    %\hline
        & \textbf{\sem{succession}} & \textbf{\sem{juxtaposition}} & \textbf{Distractors}\\
        \midrule
        \textbf{train} & 289 & 287 & 287 \\
        \textbf{test} & 731 & 678& \hphantom{0}72 \\
    \end{tabular}
    \caption{Number of noun--\textit{to}--noun sequences: two meanings of the NPN Construction, as well as \textit{distractors}. Train sets are balanced to be equal between the categories. The remaining examples are left for testing.}
    \label{tab:table_1}
\end{table}

The resulting dataset contains many instances of very common N\textit{to}N constructions, such as ``day to day''. We control for the effect of these frequent lemmas in two ways. Firstly, we artificially shrink the dataset by randomly sampling 20 sentences for each noun lemma which occurs more than 20 times, and discard the remaining sentences for the purposes of model training and testing. This is to make sure that no overly common lemmas have an overstated impact on the probing classifier performance. 

Secondly, we generate random train/test splits based on lemma of the noun in the N\textit{to}N, meaning that there are no lemmas that are seen in both the training set and the testing set. In other words, if an example with ``day to day'' is seen during training, a sentence with ``day to day'' will never be seen during testing (but a sentence with ``week to week'' might be). Each sentence in the dataset has one target instance of the N\textit{to}N construction. 

In \cref{tab:table_1}, we report the final dataset sizes, split by semantic subtype for the construction examples. N\textit{to}N \textit{constructions} are much more frequent than the N\textit{to}N \textit{distractor} patterns which serve as their near minimal pairs. We choose to balance the sizes of the two types of examples during training. We take 80 percent of the N\textit{to}N \textit{distractor} patterns for training and withhold twenty percent. We take a similar number of N\textit{to}N \textit{constructions} for training and then test on the remainder, ensuring training sets are balanced between \textit{constructions} and \textit{distractors}.

\section{Experiment 1: Constructions vs.\ Distractors}\label{sec:exp1}

\subsection{Methodology}

% \begin{figure}
%     \centering
%     \includegraphics[width=0.9\columnwidth]{images/NPN_figure1_v2.png}
%     \caption{Accuracy of N\textit{to}N \textit{construction} across layers of BERT-base. Maximal accuracy is at layer 8 (87.39\%). The black line represents control probe accuracy, which hovers around chance.}
%     \label{fig:graph_NPN1}
% \end{figure}

% We use a probing classifier to probe BERT's ability to distinguish instances of the N\textit{to}N \textit{construction} from the N\textit{to}N \textit{distractor} pattern. This follows along previous work utilizing classifiers for linguistic probing, and in particular follows \citet{Weissweiler_Hofmann_Köksal_Schütze_2022}, who use a probing classifier to study the syntax of the English Comparative Correlative construction. 

% As stated previously, some linguistic questions lend themselves to the creation of minimal pairs more than others. \citet{Weissweiler_Hofmann_Köksal_Schütze_2022} note that for the Comparative Correlative, it is not trivial to find or create minimal pairs for the construction, and resort to artificial reorderings and near-minimal pairs for their syntactic probing experiments. I find that it is extremely difficult to generate minimal pairs for N\textit{to}N sentences, because the N\textit{to}N construction occurs in multiple syntactic environments (as shown in Examples \ref{ex:first} and \ref{ex:second}), and it is not immediately clear how to automatically construct minimal pairs for N\textit{to}N which maintain overall grammaticality. 

We probe the ability for BERT to distinguish natural instances of the N\textit{to}N \textit{construction} from natural examples of the N\textit{to}N \textit{distractor} pattern. To address the issue of lexical overlap, we control for the lexical cue of the nouns in N\textit{to}N by making sure there is no overlap of nouns in the training and testing data splits, as described in \cref{sec:tt_split}. However, it is still entirely possible that the classifier learns to utilize lexical similarity of the nouns in the construction, or even other words beyond the construction. We address this by providing two baseline systems which give perspective on performance based on lexical cues: a \textit{control classifier} \citep{Hewitt_Liang_2019} and a non-contextual baseline based on GloVe embeddings \citep{pennington2014glove}. 

Control classifiers involve training new classifiers based on data where the labels are randomized and correspond deterministically to word type, ideally leading to chance performance. Following \citet{Hewitt_Liang_2019}, who deterministically assign each word a POS tag for their probing experiments, we assign a random positive or negative label deterministically based on the first noun word type in the construction. The performance of these control classifiers should be near chance, in the absence of any spurious correlations which allow the classifier to solve the task given arbitrary labels.

\begin{figure}
    \centering
    \includegraphics[width=.8\columnwidth]{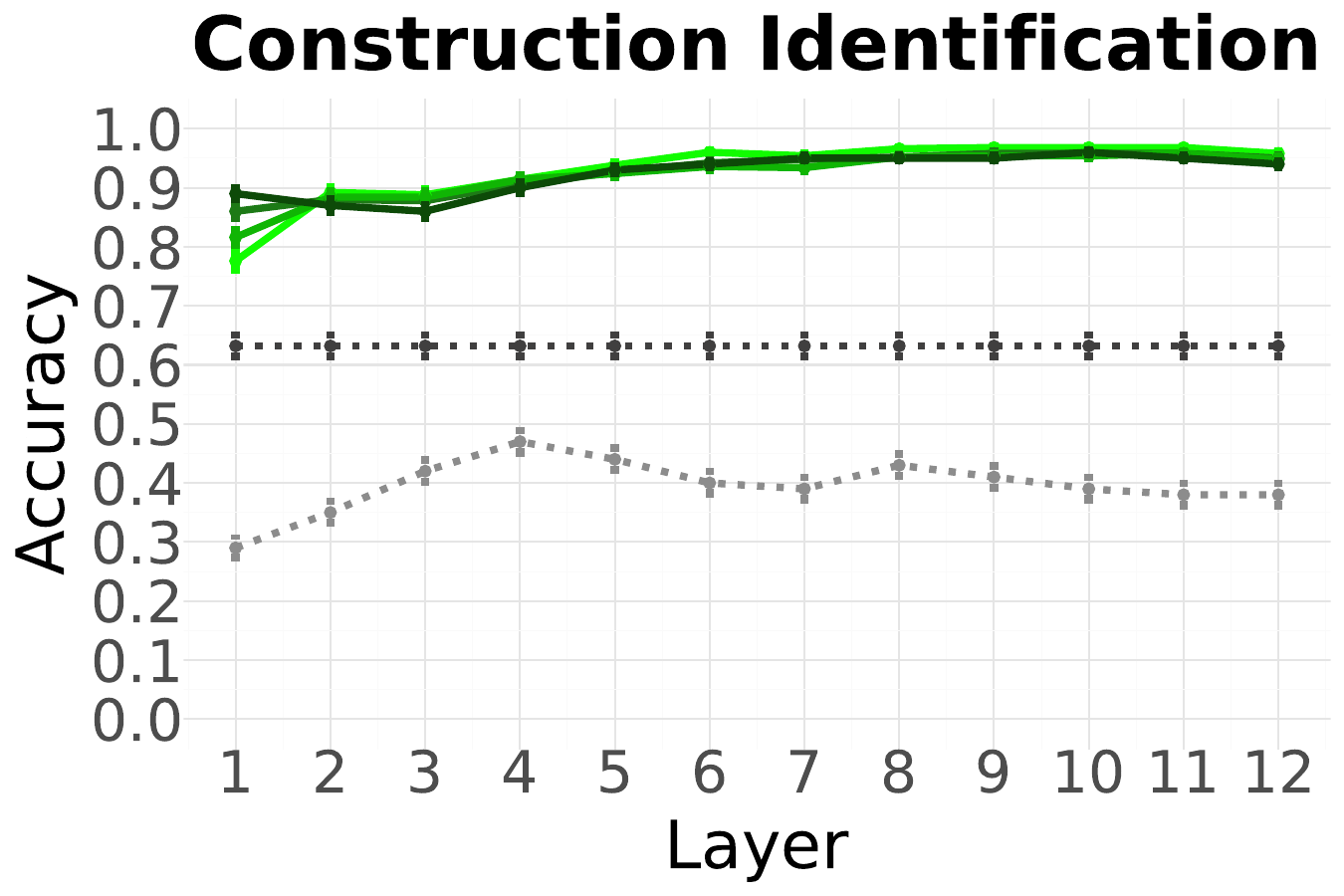}
    \caption{Accuracy of N\textit{to}N \textit{construction} \nss{identification} across layers of BERT-base, averaged across 5 random seeds \nss{as a binary? 3-way? task}. Maximal accuracy in the mid to late layers. Reducing the number of training examples does not drastically harm performance. The light grey line represents control probe \citep{Hewitt_Liang_2019} accuracy, which hovers around chance. The dark grey line represents accuracy of the lexical semantic GloVe baseline. Darker lines indicate larger amounts of training examples, with possible values of 10, 25, 100, and 287. Reducing the amount of training examples for the probes does not lead to drastically changed performance. Error Bars indicate 95\% confidence intervals over the mean accuracies across the 5 runs.}
    \label{fig:graph_NPN1}
\end{figure}

We provide an additional, non-contextual baseline by training a linear classifier on GloVe embeddings for the nouns in the construction as input. It is well known that the NPN construction is biased towards certain lexical types of nouns, such as temporal phrases and body parts \citep{Jackendoff_2008}. Thus, we expect that a classifier trained on the static embedding of the noun alone will achieve nontrivial performance. We argue that if a BERT-based classifier substantially outperforms this baseline, the difference in performance is an indication of nontrivial contextual understanding of the construction as a whole, beyond the lexical semantics of the present nouns.

Following previous probing work which tracks performance layer by layer \citet{Liu_Gardner_Belinkov_Peters_Smith_2019,Weissweiler_Hofmann_Köksal_Schütze_2022}, we train a separate probe based on embeddings from each layer of BERT and track performance across layers. We use the BERT-base-cased model, available through the Huggingface transformers library \citep{Wolf_Debut_Sanh_Chaumond_Delangue_Moi_Cistac_Rault_Louf_Funtowicz_et_al._2020}, and choose logistic regression as our linear classification architecture.\footnote{We take the embedding of ``to'' as the input into the classifier, as some past work has considered it the ``head'' of the overall construction \citep{Jackendoff_2008}.} For all experiments and data settings, we run probes with 5 random seeds and report the average results.

\subsection{Results}\label{sec:results1}

% \begin{figure}
%     \centering
%     \includegraphics[width=0.9\columnwidth]{images/NPN_figure2_v2.png}
%     \caption{Percentage of N\textit{to}N \textit{distractor} patterns that are predicted as N\textit{to}N \textit{constructions}. Lower percentages are better, since the classifier should distinguish \textit{pattern} from \textit{construction}. The best classifier is at layer 11 (7.90\% predicted as N\textit{to}N \textit{constructions}). Once again, the black dots reprsent control probe. }
%     \label{fig:graph_NPN2}
% \end{figure}

For the probing classifier results, we graph accuracy on the N\textit{to}N \textit{construction} in \cref{fig:graph_NPN1}. As we can see, the classifier is relatively strong at distinguishing the N\textit{to}N \textit{construction} from \textit{distractors} even in the early layers, with an accuracy over .90 by layer 5 with full training examples. Additionally, the classifiers are robust to sharp reductions in the number of training examples (shown in lighter shades of green in \cref{fig:graph_NPN1}), showing strong performance even with as few as 10 per-class training examples, echoing similar findings for other constructions \citep{Tayyar_Madabushi_Romain_Divjak_Milin_2020}. The control classifier achieves roughly chance performance, meaning that our trained probes have high \textit{selectivity} \citep{Hewitt_Liang_2019}. The lexical semantic baseline using GloVe achieves performance well above chance ($\approx$68\%), though its performance lags far behind the BERT-based probes, regardless of how many training example those BERT-based probes receive. This shows that overall, the probing classifier seems to be picking up on some sort of information in BERT which can reliably distinguish the N\textit{to}N \textit{construction} from its near minimal pair N\textit{to}N \textit{distractor} counterparts, beyond what is possible through lexical semantic clues alone. However, the \textit{distractor} examples generally have syntactic structure which is divergent from the \textit{construction} examples. To provide another comparison point, we now test if the existing probes can distinguish true instances of the N\textit{to}N construction from examples with artificially altered word orders.

% \begin{figure}
%     \centering
%     \includegraphics[width=0.9\columnwidth]{images/NPN_figure2_v2.png}
%     \caption{Percentage of N\textit{to}N \textit{patterns} that are predicted as N\textit{to}N \textit{constructions}. Lower percentages are better, since the classifier should distinguish \textit{pattern} from \textit{construction}. The best classifier is at layer 11 (7.90\% predicted as N\textit{to}N \textit{constructions}). Once again, the black dots reprsent control probe. }
%     \label{fig:graph_NPN2}
% \end{figure}

% \subsection{Error Analysis}

% Overall, the probing classifier is able to distinguish positive instances of the N\textit{to}N construction from the N\textit{to}N pattern distractors at an impressive accuracy. However, we do see considerable performance gains from the early layers to the later layers of BERT. Here, I do a qualitative error analysis to investigate what types of examples seem to be motivating the layer to layer performance gains. 

\begin{figure*}
    \centering
    \includegraphics[width=0.8\textwidth]{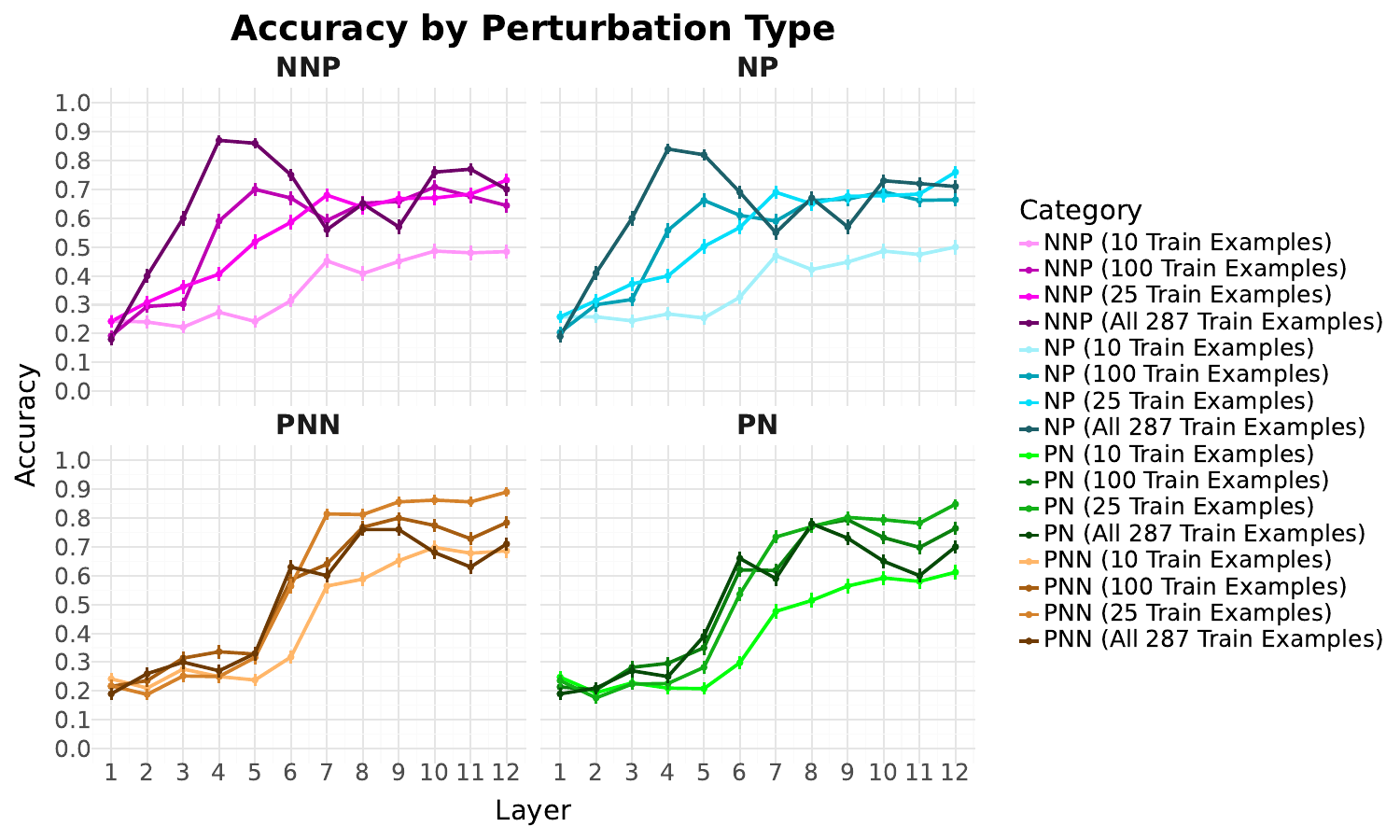}
    \caption{Accuracy of perturbed orderings of original N\textit{to}N \textit{constructions}. Since the perturbed word orders are not true instances of the construction, the true class is negative for all instances. High accuracy indicates that probes are rejecting the validity of the artificial orderings. Lighter colors represent fewer training examples for the probings. Error bars indicate 95\% confidence intervals over the average of 5 random seeds.}
    \label{fig:graph_NPN3}
\end{figure*}

\section{Experiment 2: Perturbing Word Order}\label{sec:perturb}

As we have seen in \cref{sec:results1}, a BERT-based probe can generally distinguish the N\textit{to}N \textit{distractor} patterns from the N\textit{to}N \textit{construction}. However, we wish to further test how robust the model is at distinguishing the construction from related patterns. While we have compared to naturally occuring near minimal pairs, we now test the classifier on a set of examples with artificially perturbed word order. If the classifier is robust at recognizing the N\textit{to}N \textit{construction}, it should be able to correctly distinguish \textit{construction} instances from artificial sentences with altered non-NPN word orders. To illustrate this point, consider the following two sentences:

\ex.\label{ex:nine} I need you to get this
\textbf{word for word}.

\ex.\label{ex:ten} I need you to get this
\textbf{for word word}.

Example \cref{ex:nine} is a copy of \cref{ex:second} and is a true NPN construction. On the other hand, \cref{ex:ten} is not an instance of the construction (because it does not follow the NPN word order), and is a generally ungrammatical sentence. We hypothesize that if the probe trained in \cref{sec:exp1} is not robust to the actual word order pattern of N\textit{to}N, it will be unable to distinguish sentences like \cref{ex:nine} from those like \cref{ex:ten}. If indeed the lexical cues are influencing classifier performance independent of word order, we expect that the classifier will predominantly classify examples like \cref{ex:ten} as positive instances of the N\textit{to}N \textit{construction}.

To test this hypothesis, we manipulate the test set of the probe by creating 4 perturbed orderings of each test example sentence: \textbf{\textit{PNN}}, \textbf{\textit{PN}}, \textbf{\textit{NNP}}, \textbf{\textit{NP}}. A true N\textit{to}N example is shown in \cref{ex:eleven} the corresponding 4 different perturbed orderings are shown below in \cref{ex:twelve}, \cref{ex:thirteen}, \cref{ex:fourteen}, and \cref{ex:fifteen}. 

\ex.\label{ex:eleven} Go \textbf{room to room} removing anything you don't need and selling it. (Original N\textit{to}N)

\ex.\label{ex:twelve} Go \textbf{to room room} removing anything you don't need and selling it. (PNN Perturbed Order)

\ex.\label{ex:thirteen} Go \textbf{to room} removing anything you don't need and selling it. (PN Perturbed Order)

\ex.\label{ex:fourteen} Go \textbf{room to} removing anything you don't need and selling it. (NP Perturbed Order)

\ex.\label{ex:fifteen} Go \textbf{room room to} removing anything you don't need and selling it. (NNP Perturbed Order)

Crucially, we do not retrain the linear probe on this perturbed data. This means that during training, the classifier only saw instances with the correct N + \textit{to} + N ordering, either positive instances of the N\textit{to}N \textit{construction} (like in \cref{ex:first} and \cref{ex:second}), or near minimal pairs of the N\textit{to}N \textit{distractor} patterns (like in \cref{ex:sixth}, \cref{ex:seventh}, and \cref{ex:eight}). Thus, this experiment tests the robustness of the original probing classifier when it is confronted with out of domain word orders that contain the same lexical cues as positive instances of the construction.

\subsection{Results}

% \begin{figure}
%     \centering
%     \includegraphics[width=0.9\columnwidth]{images/NPN_figure3_v2.png}
%     \caption{Percentage of perturbed orderings of N\textit{to}N \textit{distractor} patterns categorized as N\textit{to}N \textit{constructions}. Lower percentages are better.}
%     \label{fig:graph_NPN4}
% \end{figure}

% \begin{table*}
%     \centering
%     \begin{tabularx}{.84\linewidth}{|c|c|c||c|c|c|c|}
%         \hline
%          & & & \textbf{Layer 3} & \textbf{Layer 6} & \textbf{Layer 9} & \textbf{Layer 12} \\
        
%         \textbf{Lemma} & \textbf{Count} & \textbf{Gold} & \textbf{\% Correct} & \textbf{\% Correct} & \textbf{\% Correct} & \textbf{\% Correct} \\
%         \hline
%         \hline
%         \textit{Month} & 11 & Cxn & 100 & 100 & 100 & 95\\
%         \textit{Ashes} & 20 & Cxn & 0 & 5 & 0 & 0\\
%         \textit{People} & 20 & Cxn & 0 & 20 & 0 & 15\\
%         \textit{Paycheck} & 20 & Cxn & 0 & 65 & 80 & 95\\
%         \textit{Brother}  & 20 & Cxn & 45 & 25 & 30 & 20\\

%         \hline
%         \hline
%     \end{tabularx}
%     \caption{A sample of 5 lemmas and their base accurracies across layers. The Gold column is the majority Gold label for that lemma, since some lemmas occur in both the "Y" and "N" categories.}
%     \label{tab:table_2}
% \end{table*}

\cref{fig:graph_NPN3} shows the probe's performance on the perturbed test sets for the N\textit{to}N \textit{construction}. We see that in the very early layers (1--3), the probe often predicts the N\textit{to}N \textit{construction} despite the word order shifts, leading to relatively low accuracy. This possibly means that the classifier is biased by the lexical cues in the sentence early on. Interestingly, performance on \textit{PN} and \textit{PN} perturbations is substantially worse than performance on \textit{NP} and \textit{NNP} in the early layers. Accuracy on all perturbations trends upwards in the later layers, with reduction in training examples leading to drops in performance especially for \textit{NP/NNP}.

% Overall, it is an open question why the \textit{PN} and \textit{PNN} orders are generally more challenging than the \text{NP} and \text{NNP} orderings. Our hypothesis is that because in general, English prepositions precede their nouns, and thus the \textit{PN} and \textit{PNN} orderings may appear more natural than their \textit{NP} and \textit{NNP} counterparts, where the preposition follows the relevant nouns. 

%In Figure \ref{fig:graph_NPN4}, we see that perturbed ordering does not have a huge effect on the N\textit{to}N \textit{distractor} patterns, and the classifier still classifies the vast majority of these instances as non-examples of the N\textit{to}N \textit{construction}.

\subsection{Analysis}

Overall, we find that classifier probes are able to distinguish instances of the N\textit{to}N \textit{construction} from both near minimal pairs (N\textit{to}N \textit{distractor} patterns) and artificial examples (perturbed word orderings). This finding aligns with the strong performance on form-based recognition that has been observed in previous work on other constructions \citep{Li_Zhu_Thomas_Rudzicz_Xu_2022, Weissweiler_Hofmann_Köksal_Schütze_2022, Mahowald_2023}. The peak in performance in the late-middle layers is consistent with much previous work on linguistic probing, which show that the middle and late-middle layers perform best for a variety of linguistic tasks \citep{Goldberg_2019, Hewitt_Manning_2019, Lin_Tan_Frank_2019, Liu_Gardner_Belinkov_Peters_Smith_2019}. 

The differences in the performance between the \textit{NP/NNP} and the \textit{PN/PNN} perturbed orderings is an unexpected finding. According to \citet{Rogers_Kovaleva_Rumshisky_2021}, the earlier layers of BERT encode ``word order'', while the middle layers are where syntactic capabilities emerge. Based on this logic, it is unsurprising that the classifier's ability to distinguish \textit{PN/PNN} emerges in the middle and later layers. Why might the \textit{NP/NNP} instances be distinguished so much quicker? Our intuition is that in general, preposition tokens probably attend more to their immediately following word than their immediately preceding word. This is because prepositions are often immediately followed by objects, while their syntactic governor may or may not be directly adjacent to them. Perhaps in the early layers of the model (before hierarchy is as explicitly represented) prepositions attend to their following token more quickly because this is a surface word order pattern that feeds quite well into syntax.

One alternative explanation is that \textit{PN/PNN} may produce generally more grammatical sounding sentences than \textit{NP/NNP}. For instance, \cref{ex:thirteen} sounds much closer to a real sentence than \cref{ex:fifteen}. It could be that the classifier probe takes into account the ungrammaticality of \textit{NP/NNP}, even though it was not explicitly trained to do this, since the classifier probe is only trained on grammatical sentences. How exactly the ungrammaticality is represented in these embedding representations is unknown, but provides one possible explanation for the differential performance of the perturbed word ordering patterns.

Having established that performance on identifying the N\textit{to}N construction is strong, we now turn to the task of disambiguating the meaning of the construction within context.

\begin{figure*}
    \centering
    \includegraphics[width=0.9\textwidth]{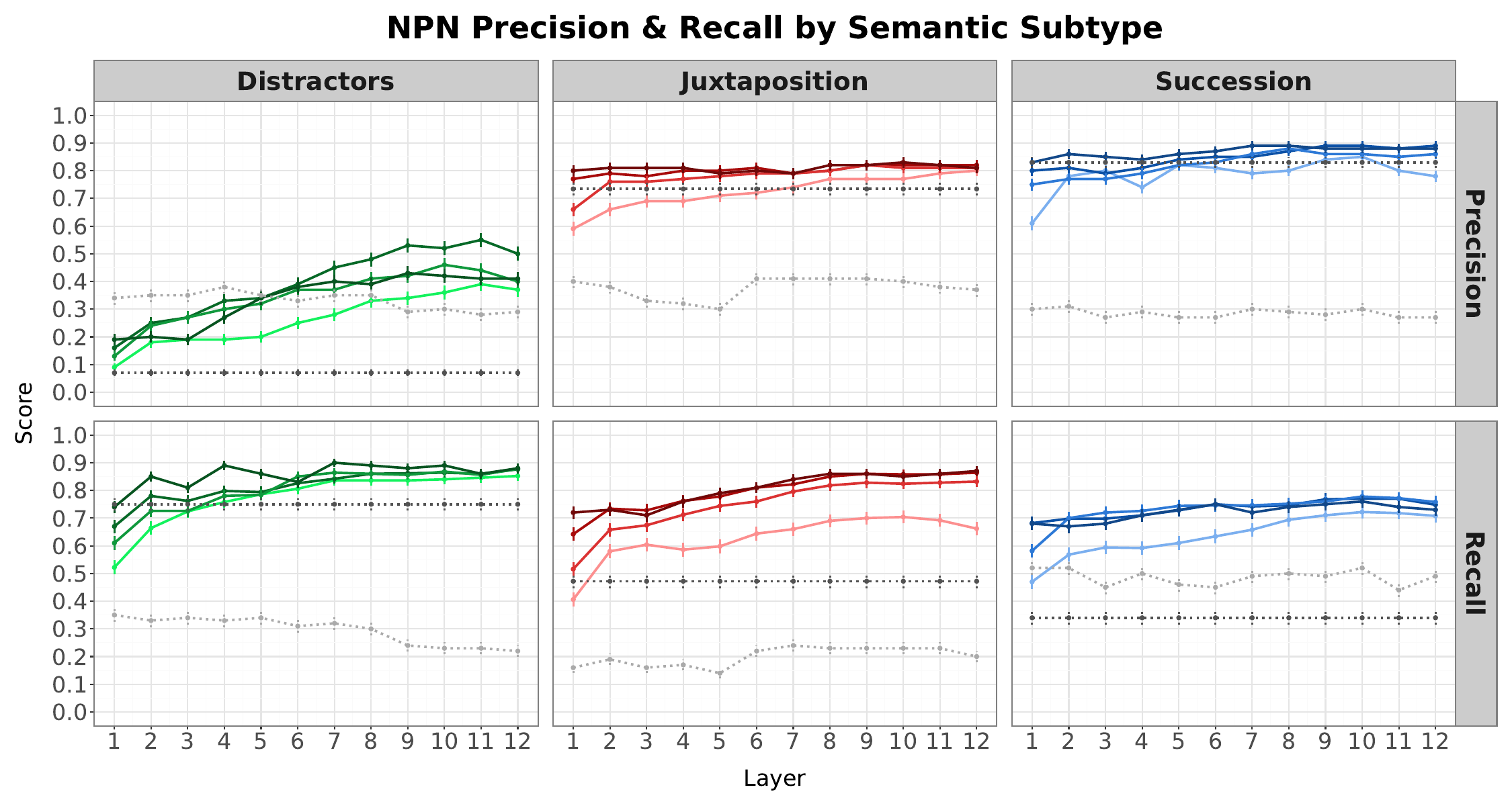}
    \caption{Precision and Recall of different semantic subtypes of NPN in 3-way classification. Lighter colors indicate fewer training examples, with possible values of 10, 25, 100, and 287 training examples per class. Classifiers trained with at least 25 per-class training examples begin to show strong performance across classes. \sem{Juxtaposition} takes substantially more training examples for classifiers to learn compared with \sem{Succession}. Each line represents the average of 5 random seeds. Dotted lines represent baselines: GloVe (black) and control (gray). Error Bars indicate 95\% confidence intervals over the average of the random seeds.}
    \label{fig:semant}
\end{figure*}

% \begin{figure*}
%     \centering
%     \includegraphics[width=0.8\textwidth]{images/line_chart_facet_prec1.pdf}
%     \caption{Precision of different semantic subtypes of NPN in 3-way classification. Classifiers trained with at least 50 training examples begin to show strong performance across classes. \textsc{Juxtaposition} takes substantially more training examples for classifiers to learn compared with \textsc{Succession}.}
%     \label{fig:semant}
% \end{figure*}

\section{Experiment 3: Semantic Disambiguation}\label{sec:exp1_semantic}

\subsection{N\textit{to}N Subtypes}
We have established that classifier performance is strong at identifying instances of the N\textit{to}N construction relative to similar patterns. However, the construction itself is ambiguous, and can have different meanings in context. The two primary meanings are \sem{succession} and \sem{juxtaposition}, which are shown in \cref{ex:third} and \cref{ex:fourth} respectively. 

The two types co-occur with different nouns at different frequencies. The \sem{succession} subtype most often occurs with spatiotemporal nouns (e.g.~\textit{day to day} or \textit{coast to coast}). On the other hand, the \sem{juxtaposition} subtype most often occurs with body parts or humans (e.g.~\textit{face to face} or \textit{friend to friend}). However, the noun meaning is not determinative, and within context some noun lemmas occur with the less common meaning. Furthermore, both constructions occur with rare noun lemmas for which it is not clear what type would be more common. 

\subsection{Methodology}
In this section, we train a classifier to distinguish semantic subtypes of N\textit{to}N. We focus on the two main subtypes that are well attested in the data: \sem{succession} and \sem{juxtaposition}. We also include examples of the N\textit{to}N \textit{distractor} patterns which are not examples of the \textit{construction}. Thus, the probe is faced with a 3-class classification problem: it must distinguish between the \sem{succession} subtype, the \sem{juxtaposition} subtype, and non-examples of the construction (\textit{distractors}). Following \citet{Hewitt_Liang_2019}, we train \textit{control classifiers} with a random label assigned to  each lemma. If the probes are properly selective, the control classifiers should have accuracies of around 33 percent.

\subsection{Results}\label{sec:results_semantic}

\cref{fig:semant} shows the precision and recall scores of the semantic probing experiments. Across all semantic types, performance is generally high for the classifiers trained on the full split of data, with recall on all 3 classes near 80\%, and strong performance even in the early layers. This is in contrast to some other semantic tasks, for which probes only reach their peaks in the mid to late layers of BERT. 

Across all layers, both \sem{succession} and \sem{juxtaposition} perform worse with only 10 training examples, but performance stabilizes after only 25 examples for the probe. The relatively low recall for \sem{juxtaposition} and \sem{succession} when the classifiers are only trained with 10 examples indicates that the probe has not fully learned to correctly distinguish the two main semantic subtypes. It is somewhat striking that there is not a larger difference between \sem{succession} and \sem{juxtaposition} in performance, given that \sem{succession} accounts for roughly 68\% of all instances of the construction in our dataset. While probes are trained with balanced training sets, the relative frequency of these semantic subtypes within our dataset (and by extension COCA) is a strong indication that \sem{succession} is the more frequent meaning. Nevertheless, performance is roughly comparable between the two semantic subtypes. In all cases, the \textit{distractor} class is overpredicted, leading to a relatively low precision compared to the subtypes of the construction. As expected, the control classifiers achieve roughly chance performance across layers, indicating that our probes have high selectivity. The GloVe-based baseline achieves an average recall of around .54 across the subtypes, but has widely variable performance depending on the semantic subtype. In general, the GloVe based classifier is much more likely to underpredict \sem{succession}, leading to very high precision and very low recall for this class.\footnote{We report GloVe and control results using the full training set. Performance of the GloVe baselines degrades with fewer examples, while the control classifiers remain near chance.}

\section{Related Work}

There has been substantial research on investigating the linguistic information that is encoded by BERT. Much of this work has focused on syntactic structure \citep{Hewitt_Manning_2019,Jawahar_Sagot_Seddah_2019,Liu_Gardner_Belinkov_Peters_Smith_2019,Hu_Gauthier_Qian_Wilcox_Levy_2020}, agreement phenomena \citep{Lin_Tan_Frank_2019} and semantics \citep{Vulić_Ponti_Litschko_Glavaš_Korhonen_2020,Chang_Chen_2019,Ettinger_2020}, with the BLiMP \citep{Warstadt_Parrish_Liu_Mohananey_Peng_Wang_Bowman_2020} and SyntaxGym \citep{Gauthier_Hu_Wilcox_Qian_Levy_2020} providing key evaluation datasets. \citet{Belinkov_2022} and \citet{Elazar_Ravfogel_Jacovi_Goldberg_2021} provide critiques of the probing classifier methodology for its indirectness and susceptibility to spurious correlations. Various improvements on the methodology have been suggested, with a general focus on providing more controlled probing environments \citep{Pimentel_Valvoda_Maudslay_Zmigrod_Williams_Cotterell_2020,Kim_Khilnani_Warstadt_Qaddoumi_2022} and causal claims through counterfactuals \citep{Ravfogel_Prasad_Linzen_Goldberg_2021,Elazar_Ravfogel_Jacovi_Goldberg_2021}. Of particular relevance to this work is \citet{Hewitt_Liang_2019}, who propose the control classifier methodology as one methodology for controlling for spurious correlations in classifier performance. We believe our use of control classifiers and non-contextual baselines provide proper context for our probing results.  

%Related to Construction Grammar, 
\nss{added:}Earlier computational linguistic work on English trained classifiers for such grammatico-semantic phenomena as identifying argument structure constructions \citep{hwang-15} and disambiguating functions of tense and definiteness \citep{reichart-10,bhatia-14}, as well as generally to disambiguate the senses of prepositions \citep{litkowski-07,schneider-18}.
\Citet{Tayyar_Madabushi_Romain_Divjak_Milin_2020} were the first to investigate BERT's performance on learning constructions, finding that BERT is able to identify a large set of hundreds of automatically identified constructions. Regarding well-established argument structure constructions, \citet{Li_Zhu_Thomas_Rudzicz_Xu_2022} find that RoBERTa implicitly contains abstract knowledge of the constructions beyond specific lexical cues. \citet{Weissweiler_Hofmann_Köksal_Schütze_2022} find that BERT-scale models are able to correctly distinguish the \sem{comparative-correlative} construction from similar looking patterns, but find that the models fail on reasoning tests related to the construction's semantics. \citet{Mahowald_2023} finds that the larger GPT-3 model can provide acceptability judgments for the Article+Adjective+Numeral+Noun (\sem{AANN}) construction which generally align with human judgements, and find that the model is sensitive to constraints on the slots in the construction. \citet{Chronis_Mahowald_Erk_2023} test BERT's knowledge of the same \sem{AANN} construction by projecting tokens in the construction into an interpretable embedding space, finding that features aligning with measure-words are evoked by tokens in the construction. Beyond BERT-scale models, \citet{Zhou_Weissweiler_He_Schütze_Mortensen_Levin_2024}, \citet{Bonial_TayyarMadabushi_2024} and \citet{Scivetti_Torgbi_Blodgett_Shichman_Hudson_Bonial_Madabushi_2025} all test LLM knowledge of constructions in more complex scenarios, finding that their performance generally lags behind humans regarding construction understanding, though there is variation depending on the construction. \citet{Zhou_Weissweiler_He_Schütze_Mortensen_Levin_2024} test a range of LLMs on understanding the \sem{causal-excess} constructions in comparison to constructions with highly similar forms, showing that the model is often misled by form-based cues. Their experiments most closely mirror our inquiries into construction sense disambiguation, though they disambiguate between similar but distinct constructions while we focus on a single polysemous 
construction. While \citet{Zhou_Weissweiler_He_Schütze_Mortensen_Levin_2024} find that LLMs largely are unsucessful at meaning-based disambiguation, and \citet{Weissweiler_Hofmann_Köksal_Schütze_2022} also find negative results regarding the semantics of the \sem{comparative-correlative}, our relatively positive results on construction disambiguation in this present work demonstrate that for N\textit{to}N, models may possess more robust models of constructional semantics than would be previously expected.  

While NPN has not been the major focus of past analysis, \citet{weissweiler-etal-2024-ucxn} do consider it as one of the constructions which they include in their UCxn dataset, which is compiled by automatically using Universal Dependencies \citep{deMarneffe_Manning_Nivre_Zeman_2021} graphs to find indications of constructions across 10 languages. We do not use this dataset due to its limited size (it contains under 50 total examples of the NPN construction in English).

\section{Conclusion}

In this work, we constructed a novel dataset of N\textit{to}N construction by extracting all instances of the construction which we found in COCA. Using our dataset, we have probed BERT's knowledge of the N\textit{to}N construction by training a linear probe to distinguish instances of the construction from near minimal pairs from corpus data. We show that a linear probe is largely able to distinguish true instances construction from naturally occurring \textit{distractor} patterns, as well as from artificially perturbed versions of the construction, though the probe is more robust to recognizing the effect of some word order changes than others. Furthermore, we show that a BERT-based classifier can disambiguate the sense of the N\textit{to}N construction in context, beyond the lexical semantic cues that are present. For both form- and meaning-based experiments, we show that the classifier results are robust even in the face of dramatic reductions in the number of training examples. This indicates that constructional knowledge is likely latently encoded within BERT and not due to spurious correlations learned by the classifiers. Overall, these results contribute to the growing body of evidence that LMs have some ability to acquire grammatical properties of rare and idiosyncratic constructions.

\section{Limitations}

This work is limited in several ways. Due to natural relative frequencies of various constructions, the dataset used for N\textit{to}N is unbalanced between the N\textit{to}N \textit{construction} and \textit{pattern}. This means that the training set for the classifier was quite small, because we ensured that training was balanced between the different classes. While the probing classifiers do achieve high accuracy, it is unclear how much accuracy is being capped by the limited data available. However, this fact, alongside our experiments with reduced training set sizes, indicate that the probes can learn with relatively little training signal.

This is experiment is also limited in only considering N\textit{to}N, as opposed to the broader NPN construction. This is an intentional choice, as ``to'' has the most semantic subtypes of NPN associated with it. Future work is needed to see if the results here are robust to the inclusion of additional NPN examples with other lemmas into the dataset. We also only consider the English NPN construction, though the construction has been observed in a range of languages, including Dutch, English, French, German, Norwegian, Japanese, Mandarin, Polish, and Spanish \citep{weissweiler-etal-2024-ucxn}. We also limit our experiments to cases where the nouns match. This choice greatly simplifies our process of detecting true constructions as well as distractors, but also excludes some interesting examples of the construction, as pointed out by \citet{Jackendoff_2008}. 

Finally, this work utilizes the probing classifier methodology, which has been criticized for providing indirect/correlational evidence of linguistic information in LM representations \citep{Belinkov_2022}. Future work is needed to broaden the analysis to include causal probing methodologies (e.g.~AlterRep, \citealt{Ravfogel_Prasad_Linzen_Goldberg_2021}; MaPP, \citealt{Karidi_Zhou_Schneider_Abend_Srikumar_2021}; Reconstruction Probing, \citealt{Kim_Khilnani_Warstadt_Qaddoumi_2022}).

\section*{Acknowledgments}

We thank CoNLL reviewers and NERT lab members for their thoughtful comments. This research was supported in part by NSF award IIS-2144881.

\bibliography{acl_latex}

\end{document}